\documentclass[fleqn,10pt]{wlscirep}
\usepackage[utf8]{inputenc}
\usepackage[T1]{fontenc}
\usepackage{comment}
\usepackage[linesnumbered,lined,boxruled,commentsnumbered]{algorithm2e}
\usepackage{geometry}
\usepackage[noend]{algpseudocode}
\usepackage{subcaption}
\usepackage{wrapfig}
\usepackage{amsmath}
\usepackage[export]{adjustbox}
\usepackage{caption}
\usepackage{environ}
\usepackage{array}

\NewEnviron{HUGE}{% 
    \scalebox{1.25}{$\BODY$} 
}

\usepackage{multirow}
\title{Dynamic Zoning of Industrial Environments with Autonomous Mobile Robots}

\author[1,*]{Russell Keith}
\author[1,*]{Hung La}
\affil[1]{University of Nevada-Reno, Computer Science and Engineering, Reno, NV 89503, United States}
\affil[*]{corresponding.keithr@unr.edu, hla@unr.edu}

\begin{abstract}
This paper presents a scheduling algorithm that divides a manufacturing/warehouse floor into zones that an Autonomous Mobile Robot (AMR) will occupy and complete part pick-up and drop-off tasks. Each zone is balanced so that each AMR will share each task equally. These zones change over time to accommodate fluctuations in production and to avoid overloading an AMR with tasks. A decentralized dynamic zoning (DDZ) algorithm is introduced to find the optimal zone design, eliminating the possibility of single-point failure from a centralized unit. Then a simulation is built comparing the adaptability of DDZ and other dynamic zoning algorithms from previous works. Initial results show that DDZ has a much lower throughput than other dynamic zoning algorithms but DDZ can achieve a better distribution of tasks. Initial results show that DDZ had a lower standard deviation of AMR total travel distance which was 2874.7 feet less than previous works. This 68.7\% decrease in standard deviation suggests that AMRs under DDZ travel a similar distance during production. This could be useful for real-world applications by making it easier to design charging and maintenance schedules without much downtime. Video demonstration of the system working can be seen here: \url{https://youtu.be/yVi026oVD7U} 
\end{abstract}
\begin{document}

\flushbottom
\maketitle
\thispagestyle{empty}

%\noindent Please note: Abbreviations should be introduced at the first mention in the main text – no abbreviations lists. Suggested structure of main text (not enforced) is provided below.

\section*{Introduction}

\begin{wrapfigure}[15]{r}{0.5\textwidth}
    \centering
    \includegraphics[width=2.7 in]{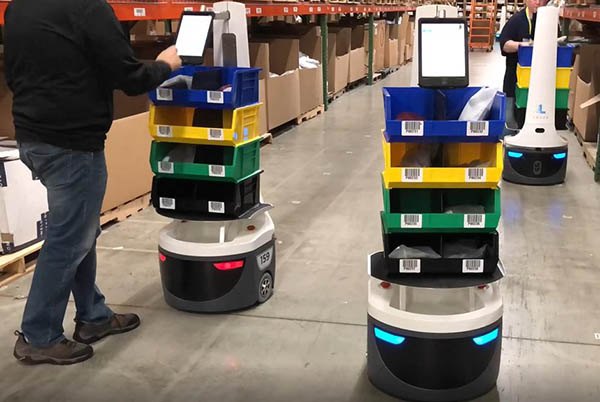}
    \caption{Human-robot collaboration in the warehouse~\cite{hart_efficient_2023}}
    \label{Fig: 1a}
\end{wrapfigure}

 With recent hardware advances in sensors and computing power, autonomous mobile robots (AMRs) have become more feasible in manufacturing and warehouse environments. For companies to stay competitive and become more Lean~\cite{longhan_production_2013}, many have turned to an automated production process that deploys a fleet of AMRs or automated guided vehicles (AGVs) to complete simple pick-up and drop-off tasks (see figure ~\ref{Fig: 1a}). This alleviates workers from repetitive tasks and frees up time for more valuable production.
\begin{comment}
Automated or "smart warehouses" do not just use AMRs, they also deploy a series of tools for better planning strategies, resource management, and path planning. A good example is how Amazon utilizes their Kiva robots in their parts-to-picker system~\cite{banker_new_nodate}. Other companies, such as Matthews Automation~\cite{noauthor_autonomous_nodate}, offer AMRs in conjunction with warehouse management software to optimize production and control the fleet.

While large companies have enough capital to invest in automation, smaller companies have to consider other factors such as the high initial investment of an AMR fleet (\$150 - \$200 per square foot). A fleet can also have an impact on the general flow of parts especially if switching picker-to-parts or part-to-picker systems. Even though automation comes at a high cost, the returns from reducing production waste have shown to be substantial in the longer term~\cite{kamali_smart_2019}.
\end{comment}

With the introduction of AGVs and AMRs to industrial environments, many scheduling algorithms have been developed to manage orders while they are moving through the floor. Many of these algorithms can find the optimal order schedule, however, most are tailored toward the warehouse environment. Unlike warehouses, manufacturing environments are not as predictable since there are many different types of parts going to different assemblies and machines. Often, there are daily inconsistencies with production, which are caused by temporary delays like broken tools, missing parts, repair, and personnel. 

One way to approach scheduling in manufacturing environments is by sectioning off areas of the warehouse floor into zones where a robot will operate. Each robot performs pick-up and drop-off tasks from one point to another, assuming that a human will be at the same point to physically manage the items. The system will continually monitor and record data to determine if there is an imbalance between zones. If an imbalance is detected and persists, it will be corrected by redrawing zones based on the current data. 

Ho and Liao~\cite{ho_zone_2009} introduced a simulated annealing (SA) algorithm with dynamic zoning for use with AGVs. This paper builds upon that algorithm and adjusts it for use with AMRs. For comparison, a genetic algorithm (GA) is also implemented replacing SA from the dynamic zoning algorithm in Ho and Liao~\cite{ho_zone_2009}. We also introduce a Decentralized Dynamic Zoning (DDZ) algorithm that focuses on designing zones around the average load between each robot. DDZ uses Weighted Average Consensus~\cite{xiao_scheme_2005} to find the average load in the network and then adjusts zones to lower the load standard deviation by independently running SA between each robot in the network. The system assumes that robots have a limited communication range and can only build zones based on their neighbor's information. Zone information is shared with workstations that push parts to the next robot once they are finished processing.  
 \begin{table}[t]
    \centering
    \begin{center}
    \begin{adjustbox}{max width=\linewidth}
      \begin{tabular}{m{1.25in}m{1.5in}m{1.5in}m{1.5in}m{1.5in}m{1.5in}m{.1in}}
        \hline
        \textbf{Method} & \textbf{Scheduling for use with multiple AMRs} & \textbf{Adjusts to a dynamic work environment} & \textbf{Used with a manufacturing environment} & \textbf{Used with a warehouse environment} & \textbf{Decentralized}\\
        \hline
        Simulated annealing\cite{ho_zone_2009} & \boxed{} & \checkmark & \checkmark & \boxed{} & \boxed{}\\\
        Min-Max Strategy\cite{yokota_min-max-strategy-based_2019} & \boxed{} & \boxed{} & \boxed{} & \checkmark & \boxed{}\\
        Two-stage heuristic\cite{yu_multi-load_2021} & \boxed{} & \boxed{} & \boxed{} & \checkmark & \boxed{}\\
        SAC and reinforcement learning\cite{tang_novel_2021} & \boxed{} & \checkmark & \boxed{} & \checkmark & \boxed{}\\
        Dynamic task chain\cite{li_research_2022} & \boxed{} & \checkmark & \boxed{} & \checkmark & \boxed{}\\
        Auction based method\cite{basile_auction-based_2017} & \checkmark & \checkmark & \checkmark & \checkmark & \boxed{}\\
        GA Dynamic zoning & \checkmark & \checkmark & \checkmark & \checkmark & \boxed{}\\
        DDZ & \checkmark & \checkmark & \checkmark & \checkmark & \checkmark\\
        \hline
      \end{tabular}
    \end{adjustbox}
    \end{center}
    \caption{Scheduling algorithm comparison: Addressed \checkmark, Not addressed \boxed{}}
    \label{schduleing comp}
\end{table}
 Table~\ref{schduleing comp} compares the method proposed in this paper and that of other research. The proposed method has the advantage of being applicable in both warehouse and manufacturing facilities. This method is designed in unpredictable environments where robots are only informed of pick-up and drop-off locations. DDZ relies on finding a load balance among all robots so that tasks can be equally shared between each robot. The main contributions of this paper are as follows:

\begin{itemize}
    %\item Introduce a GA that can heuristically find the best zone partition design. This algorithm is able to explore a solution space by freely adding or removing workstations that are adjacent to the zone. The best zone partition design is kept and used in simulation.
    \item Introduce a DDZ algorithm that finds the best zone design by focusing on reducing the standard deviation between the robot load and the calculated average. Each robot can only communicate with the robots that are within communication range and the workstations within their zone. 
    \item  Compare DDZ to GA and previous works with SA~\cite{ho_zone_2009} using a realistic simulated manufacturing floor. Throughput and distance traveled are compared to highlight the advantages and disadvantages of each method.
    \item Provide the community with open source code for AMR experimentation and for use as a platform to build manufacturing/warehouse simulations using NVIDIA ISAAC Sim~\cite{nvidia_developer_isaac_2024}.
\end{itemize}

\section*{Methods}
\subsection*{Literature Review}
%Topical subheadings are allowed. Authors must ensure that their Methods section includes adequate experimental and characterization data necessary for others in the field to reproduce their work.
Warehouse and manufacturing scheduling is considered to be an NP-hard problem ~\cite{anghinolfi_bi-objective_2021}, which can be solved with heuristic approaches such as GAs, reinforcement learning, or SA. A lot of research has gone into warehouse scheduling like that of Yokota~\cite{yokota_min-max-strategy-based_2019}. In Yokota~\cite{yokota_min-max-strategy-based_2019}, the authors focus on the coordination of AGVs with human pickers, having the robots meet with humans at the pick location and the human picker physically placing the item on the AVG. The algorithm works by first assigning orders as equally as possible to each of the AGVs and then optimizing the route with SA. Then, a human is assigned to meet an AGV based on the proximity and current burden of the picker. Other papers consider robot resources such as battery consumption. In Li and Wu~\cite{li_research_2022}, the authors propose a Genetic Algorithm Considering Genome algorithm that finds the best task and charging schedule for multiple AGVs. It aims to reduce the amount of competition between robots and charging stations, assuming that there are more robots than there are stations. 

Though there are many scheduling algorithms made for warehouse environments, there have been few scheduling algorithms made for manufacturing environments. The facilities differ in their approach to scheduling in that in manufacturing, assembly workers are in a single area and robots move parts from station to station rather than from an aisle to a common drop-off area. The manufacturing floor is also more unpredictable since machinery and tools are constantly moving through aisles and blocking paths. Other factors like part processing time and workstation queue size also need to be considered when scheduling tasks. In a paper, Rizvan \emph{et al.}~\cite{erol_multi-agent_2012} uses a multi-agent approach to address scheduling issues in a dynamic environment. The algorithm works in real time and uses a bidding system between agents for tasks. Each AGV bids on how fast it can complete a task and once the best bid has been accepted, the task is then rewarded to the corresponding AGV. This system schedules not only AGVs but the machines in the system as well. A simpler approach to manufacturing scheduling is with dynamic zoning. Ho and Liao~\cite{ho_zone_2009} propose an SA method to find the best way to divide a manufacturing floor into zones so that every AGV operating in each zone will have a balanced workload. 

To the author's knowledge, there have only been a few studies that focus on decentralized scheduling. Hierarchical control algorithms often find the optimal solution but suffer from a lack of robustness or flexibility. In these control algorithms, if the central control unit falters the entire system cannot operate. Decentralized control has the advantage of scalability and continuous operation even if one unit fails. Basile \emph{et al.}~\cite{basile_auction-based_2017}, used a version of an auction-based approach called the sealed-bid method in which an agent cannot see other agents’ bids. Any robot can play the role of auctioneer and take in bids from other robots acting as agents. The auctioneer selects the best bid based on the time it takes to complete a task. In another paper by Warita and Fujita~\cite{warita_online_2024}, the authors set up a multi-agent decentralized algorithm using a fully decoupled upper confidence tree (FDUCT). The idea is that each AGV will use FDUCT to plan its actions while predicting the actions of other agents. They also implement an item-exchange strategy to help balance the load of each AGV and avoid idle time. 

\subsection*{Concepts}

\begin{table}[!th]
    \centering
    \begin{adjustbox}{max width=\linewidth}
    \begin{tabular}{p{.50in}p{3.5in}p{.50in}p{3.5in}}\hline
        
        $V$ & Velocity of robot  &  $M$ & Number of temperature reductions  \\
        
        $nz$ & Number of zones &  $n$ & Iteration\\
        
        $L_{pz}$ & Load of a vehicle in zone $z$ of a zone partition design $p$ (min) & $k$ & Weight temperature constant\\
        
        $DA_{pzij}$ & The total distance of empty trips from workstation $i$ to $j$ in zone $z$ of design $p$ & $x_i$ & Robot $i$ average load measurement, initialized to current robot load \\
        
        $DB_{pzij}$ & The total distance of loaded trips from workstation $i$ to $j$ in zone $z$ of design $p$ &  $N_i$ & Set of robot $i$ neighbors, defined by a radius from robot $i$ \\
        
        $g_{pzij}$ & The expected number of empty trips from  workstation $i$ to $j$ in zone $z$ of design $p$ &  $\xi$ & Edge set in communication graph\\

        $f_{pzij}$ & The number of loaded trips from  workstation $i$ to $j$ in zone $z$ of design $p$ & $W_{ij}$ & Linear weight on $x_j$ at robot $i$ \\
        
        $d_{pzij}$ & The path distance between workstations $i$ to $j$ in zone $z$ of design $p$ & $d_i$ & Number of neighbors of robot $i$ \\
      
        $ws(p,z)$ & Set of workstations in zone $z$ of design $p$ & $q_i$ & Position of robot $i$ \\
        
        $\sigma$ & Standard deviation as seen from robot $i$ and calculated with $x_i$ &  $r$& Robot communication range \\
        
        $Li$ & Robot $i$ load & $ts$ & Transfer stations\\
        
        $T_c$ & Current temperature & $T_{AC}$&  Time period to run Weighted Average Consensus\\
        
        $T_i$ & Initial temperature & $L_{TOL}$ & The amount of load difference that is tolerable between $X_i$ and $L_i$\\

        $T_f$ & Freezing temperature & $T_{LT}$ & Time limit to declare an imbalance and to start DDZ \\

    \hline
    \end{tabular}
    \end{adjustbox}
    \caption{Notation~\cite{ho_zone_2009}}
    \label{notation}
\end{table}

\vspace{-15pt}
\subsubsection*{Zones}

A zone is a series of critical points and segments that connect workstations (WS). Critical points are points on the floor where two aisles meet or where a workstation is located. Critical segments are the pathways connecting two critical points. For example, in figure~\ref{ex floor layout}, G, H, and I are critical points and the lines between them $\overline{GHI}$ are the critical segments connecting them. A zone is a collection of critical segments connecting workstation to workstation. Zone 1, in figure~\ref{ex floor layout}, has three workstations with critical segments $\overline{HINO}$,$\overline{ONSR}$ connecting \textit{WS1}, \textit{WS4}, and \textit{WS5}.

\subsubsection*{Shortest Feasible Path and Adjacency}
The shortest feasible path is a collection of critical segments that connect one workstation to another. This path must satisfy two conditions. The first is that the two workstations that are connected by a path, must belong to the same or two different zones. The second is that the path does not go through another zone other than the zone or zones that either workstation belongs to.
A workstation is considered to be adjacent to another workstation if it is within a certain Manhattan distance. Adjacency does not consider zones and is only a distance comparison. Adjacency distance is dependent on the layout of the floor.

\subsubsection*{Tip Workstations}
\begin{wrapfigure}[14]{r}{0.5\textwidth}
%\begin{figure}
    \vspace{-15px}
    \centering
    \includegraphics[width=3.55in]{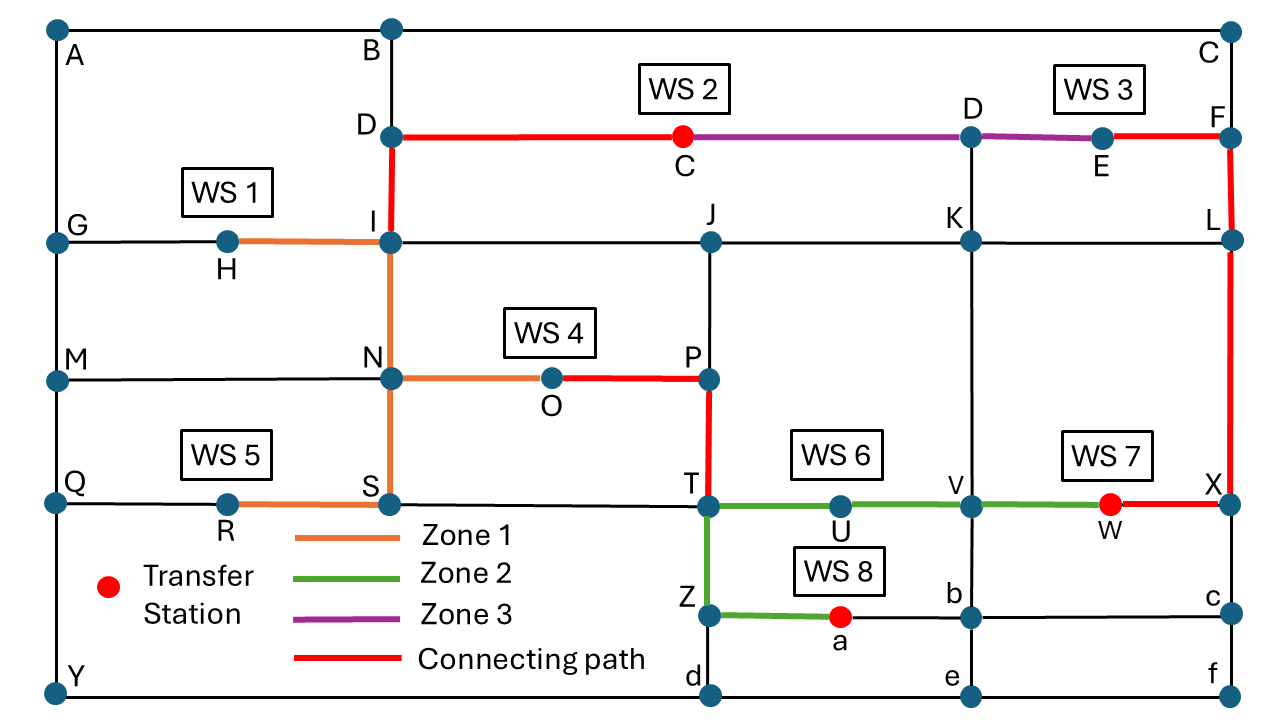}
    \caption{Example zone design without transfer stations}
    \label{ex floor layout}
    \vspace{-15px}
%\end{figure}
\end{wrapfigure}
A tip workstation is defined to be any workstation with only one branch or one series of critical segments connecting it to the rest of the workstations in the same zone. In figure \ref{ex floor layout}, zone 1 has three tip workstations WS1, WS4, and WS5. However, zone 2 only has two tip workstations because WS6 has two critical segments connecting to it. When a tip workstation is removed from a zone, the critical segments connecting it to the zone are also removed. For example, if WS4 were to be removed from zone 1, the zone would go from: (WS2, WS4, WS5), $\overline{HINO}$,$\overline{ONSR}$ to: (WS2, WS5), $\overline{HINSR}$ (see figure \ref{ex floor layout}).

\subsubsection*{Transfer Stations}
Zones are occupied by one robot that is responsible for the pick up and drop off of parts within that zone. If a part needs to go into a different zone, then a robot will drop it off at a transfer station where the robot in the next zone will pick it up and take it to its next location. Transfer stations are workstations that are close to an adjacent zone and have a path connecting them to a workstation in the adjacent zone. For DDZ, the connecting paths from the zone to the transfer station can cross each other but do not cross any other primary zones. The primary zones are the critical segments connecting workstation to workstation not including workstation to transfer station. Figure~\ref{ex floor layout} shows how transfer stations can connect zones. Zone 1 and zone 2 share WS2 as a transfer station with a connecting path $\overline{HIDC}$ assigned to zone 1. Algorithm~\ref{find_ts} describes the how to find transfer stations between two zones. 

\begin{minipage}[t]{0.48\textwidth}
    \begin{algorithm}[H]
    \caption{Finding transfer stations between two zones}\label{find_ts}
    \SetAlgoLined
        \SetKwInOut{Input}{Input}
        \SetKwInOut{Output}{Output}
        \Input{zone A (A), zone B (B)}
        \Output{Set of transfer stations between A and B ($TS_{AB}$)}
        $A_{\textit{tip}} \gets$ set of tip workstations in $A$\\
        $B_{\textit{tip}} \gets$ set of tip workstations in $B$\\
        $TW_A \gets$ set of $B_{\textit{tip}}$ that are adjacent to $A_{\textit{tip}}$\\
        $TW_B \gets$ set of $A_{\textit{tip}}$ that are adjacent to $B_{\textit{tip}}$\\
        \textit{ABP}$\gets$set of critical segments that belong to A and B and those not assigned to a zone\\
        $SPC \gets$ set of shortest feasible paths between every workstation in $TW_A$ and $TW_B$ using ABP\\
        \For{$WS_\alpha \in SPC$}{
            $CSP \gets$ shortest path in SPC leading to a tip WS in the other zone\\
            $WS_\beta \gets$ connecting tip WS in $CSP$\\
            $\alpha \gets $zone that claims $WS_\alpha$\\
            $\beta \gets $zone that claims $WS_\beta$\\
            $L_\alpha \gets$ load in zone A\\
            $L_\beta \gets$ load in zone B\\
            \If{$L_\alpha >= L_\beta$}{
                \textit{TS}$_{\alpha\beta}\gets WS_\alpha$
                $\beta\gets CSP$\\
                \If{$L_\alpha < L_\beta$}{
                \textit{TS}$_{\alpha\beta}\gets WS_\beta$\\
                $\alpha\gets CSP$
                }
            }
            remove $CSP$, $WS_\beta$, $WS_\alpha$ from SPC
        }
        \textbf{return} $TS_{AB}$
    \end{algorithm}
\end{minipage}
\begin{minipage}[t]{0.48\textwidth}
    \begin{algorithm}[H]
    \caption{Shortest job first with aging}\label{shortest job}
    \SetAlgoLined
        \SetKwInOut{Input}{Input}
        \SetKwInOut{Output}{Output}
        \SetKwProg{Fn}{Function}{ :}{end}
        \Input{parts in robots queue ($R_L$), position of robot ($R_{pos}$), velocity of robot ($V$)}
        \Output{sorted queue of parts ($S_p$), next part to pick up ($P$)}
        \Fn{dist($A$, $B$)}{
            $AB_{path} \gets$ find the shortest path from point A to B\\
            $AB_{dist} \gets$ total distance of $AB_{path}$\\
        \Return $AB_{dist}$
        }
        $C_D \gets $weight of distance score\\
        $C_A \gets $weight of age score\\
        \For{part ($P$) in $R_L$}{
            $P_p \gets$ $P$ pick-up location\\
            $P_d \gets$ $P$ drop-off location\\
            $jobdist \gets dist(R_{pos},P_p) + dist(P_p,P_d) $\\
            $P_{age} \gets$ time part has been in queue\\
            $S_p \gets \{C_A(P_{age}) - C_D(jobdist/V))$: $P$\}\\
            }
        sort $S_p$ from highest to lowest score (key)\\
        $P \gets$ robot's next part is the highest scoring part in $S_p$\\
        \textbf{return} $S_P, P$
    \end{algorithm}
\end{minipage}

\subsubsection*{Shortest Job First with Aging}
%\begin{wrapfigure}[14]{r}{0.5\textwidth}
%\begin{figure}
    %\vspace{-15px}
    %\centering
    %\includegraphics[width=3.55in]{images/CustomFig1.png}
    %\caption{Example zone design without transfer stations}
    %\label{ex floor layout}
    %\vspace{-15px}
%\end{figure}
%\end{wrapfigure}
Once zones are established, and robots have several parts in their queue, a method is needed to organize the part queue. There are many different approaches to scheduling tasks with a common approach being "shortest job first". This method is often used with CPU processing and can be mathematically proven to have the shortest average waiting time~\cite{modern_operating_systmes_2015}.

"Shortest job first" prioritizes tasks based on the shortest completion time. Though this method is simple and not computationally taxing, a downside to using this method is that it can lead to longer tasks being ignored if a robot's queue is filled with short-distance parts. To compensate, an aging factor is introduced to prioritize parts on their shortest completion time, and how long they have been waiting in queue. Algorithm~\ref{shortest job} shows how tasks are scored and ranked in the robot queue. $C_A$ and $C_B$ are weight factors that offset job completion time vs. age. Variable $dist(A,B)$ is the shortest Manhattan path from WS A to WS B ignoring zone design, since this is the path the AMR will take. Aging starts when a part has finished processing and is reset after processing at the next workstation. Age is preserved while moving through transfer stations.

\subsubsection*{Zone Load}
Load is a measure of how much time a robot will spend moving parts within a zone. Equations (\ref{L_{pz}})-(\ref{g_{pzij}})\cite{ho_zone_2009} show how to calculate load. $f_{pzij}$ captures the total flow from $i$ to $j$. $f_{pzij}$ includes not only the flow of parts from $i$ to $j$ but also transferred parts from other zones that pass through $i$ and arrive at $j$, and parts that pass through $i$ and $j$ and arrive at different workstations in the other zones. For DDZ, zone load for robot $i$ ($L_i$) is calculated by equation~\ref{L_{pz}}, zone \textit{z} includes not only the workstations in the primary zone but also the transfer stations between each neighbor. Flow is only calculated from workstations in \textit{z} since each robot is only aware of the parts that are in its queue. This means that robots only know where to pick up and drop off parts next, not necessarily the starting pick-up and final drop-off locations where parts will be processed. If the robots only considered workstations that are in their primary zone, then parts that are at, or going to transfer stations will be missed when calculating zone load. See Ho and Liao~\cite{ho_zone_2009} for a complete breakdown of equations.

%\begin{minipage}{0.45\linewidth}
    %\begin{gather}\label{SV_p}
        %SV_p = \frac{TZLD_p}{\sum_{z'=1}^{nz}{L_{pz'}} \times (nz - 1)}
    %\end{gather}
    %\begin{gather}\label{TZLD_p}
        %TZLD_p = \sum_{z'=1}^{nz-1}\sum_{z''=z'+1}^{nz}{\lvert L_{pz'}-L_{pz''}\rvert}
    %\end{gather}
    \begin{equation}\label{L_{pz}}
        L_{pz} = \frac{\sum_{i\in ws(p,z)}\sum_{j\in ws(p,z)}(DA_{pzij}+DB_{pzij})}{V} + (\sum_{i\in ws(p,z)}\sum_{j\in ws(p,z)}f_{pzij})(t_u+t_l)
    \end{equation}
    \begin{equation}\label{DA_{pzij}}
        DA_{pzij} = g_{pzij}d_{pzij}
    \end{equation}
%\end{minipage}
%\begin{minipage}{0.45\linewidth}
    \begin{equation}\label{DB_{pzij}}
        DB_{pzij} = f_{pzij}d_{pzij}
    \end{equation}
    \begin{equation}\label{g_{pzij}}
        g_{pzij} = \frac{\sum_{k\in ws(p,z)}f_{pzki}\sum_{k\in ws(p,z)}f_{pzjk}}{\sum_{m\in ws(p,z)}\sum_{n\in ws(p,z)}f_{pzmn}}
    \end{equation}

\subsection*{Method: Decentralized Dynamic Zoning (DDZ)}

\subsubsection*{Introduction}
In this section, we will discuss a decentralized dynamic zoning algorithm and the benefits this approach has over centralized algorithms from previous works. The section describes an algorithm that dynamically adjusts zones over time based on the calculated average load. The Results section of this paper contains the experimental outcome of this algorithm.

%To this author's knowledge, there are not many decentralized scheduling algorithms that are capable of responding to changes in workflow in real time. Since robots can come to conclusions by themselves, decentralized algorithms have the advantage of being more robust allowing for easier adding or subtracting of robots.

Algorithms from previous works (GA and SA from Ho and Liao~\cite{ho_zone_2009}) find the optimal zone design by balancing the load of each robot. Load is an expression of how much time a robot will take to pick up and drop off all parts that have been assigned to it. It is calculated through parts that are currently making their way through the system and what workstation each part has visited, also known as part history. Though the intention is to share the load of overloaded robots with under-loaded robots, an inefficient zone design that splits zones in a way that extends part travel through several zones could be chosen if the loads of each zone are similar. Including part history in the load calculation also comes with a downside, if a zone ceases to see a high-frequency part, the zone load will remain high until the part history is cleared. This leads to sluggish reactions to changes in part load.

A potentially better approach could be to design zones that try to reduce the error between the robot load and the average robot load. The load is calculated based on what each robot sees in its current queue and not on the part history. With this, zones have more of an incentive to distribute parts among other robots leading to reduced travel time variation. The focus of the optimal zone design will also center around what each robot currently sees in its queue. Responding to what the current load is rather than what the future load might be.

\subsubsection*{Decentralized Zoning}

Unlike the GA and SA~\cite{ho_zone_2009} algorithms, the robots using the decentralized dynamic zoning (DDZ) are responsible for finding the best zone design independently. To achieve this, this paper makes the following assumptions.

\begin{itemize}
    \item Robots can only communicate with other robots within a certain communication range.
    \item Workstations are independent and only communicate with their claimed robot(s).
    \item Workstations are assumed to have a much larger communication range than robots and can communicate with any robot that claims it as a workstation or transfer station. 
    \item Robots are only aware of parts that are in or going through their zone.
    \item Parts are passed between two robots through transfer stations only.
    \item If a robot does not have any neighbors within communication range during zone redesign then it will continue to move until it has at least one neighbor.
\end{itemize}

The algorithm works by first finding the average robot load using Weighted Average Consensus with Metropolis Weights~\cite{xiao_scheme_2005, La_IEEE_C2013} (algorithm~\ref{WAC}). Each robot will then compare its current load with that of the average. If the robot's load is out of tolerance ($L_{TOL}$) for $T_{LT}$ time then the robot will signal to start DDZ. This signal is sent to neighboring robots and they, in turn, signal to their neighboring robots to start DDZ. The first robot to signal will then be designated as the leader and algorithm~\ref{DD_zoning} will commence. The algorithm uses SA between each robot to lower the standard deviation of load (equation~\ref{DDZstd}) until each robot reaches its freezing temperature. To make a better comparison between the SA algorithm from Ho and Liao~\cite{ho_zone_2009} and DDZ, we will use the same temperature reduction strategy. However, since our solution space is in the range $[-\infty,\infty]$ a constant $k$ is included in the probability calculation to weigh the temperature appropriately. Equation~\ref{DDZtemp} shows the temperature strategy~\cite{ho_machine_1998}. Algorithm~\ref{DD_zoning} further describes
how DDZ finds the optimal zone design.

\begin{wrapfigure}[21]{l}{0.5\textwidth}
    \vspace{-20px}
    \begin{minipage}{\linewidth}
        \begin{gather}\label{DDZstd}
             \sigma = \sqrt{\frac{(L_i+x_i)^2 + \sum_{j \in Ni}{(L_j-x_i)}^2}{di + 1}}
        \end{gather}
        \begin{gather}\label{DDZtemp}
             T_c = T_i(\sqrt[M]{\frac{T_f}{T_i}})^{n}
        \end{gather}
        \begin{algorithm}[H]
        \caption{Weighted Average Consensus with Metropolis Weights}\label{WAC}
            \begin{algorithmic}
                \State $x_i(t+1)=W_{ii}(t)x_i(t)+\sum_{j \in N_i(t)}W_{ij}(t)x_j(t)$, $i = 1, ...,nv$
                \State \[ W_{ij} = \begin{cases}
                    \frac{1}{1+\max\{d_i(t),d_j(t)\}},& \text{if} \{i,j\}\in\xi(t)\\
                    1-\sum_{\{i,k\}\in\xi(t)}W_{ik}(t), & i=j\\
                    0, & \textit{otherwise}
                \end{cases}
                \]
                \State $N_i(t) = \{j \in \delta:\|q_j-q_i\| \leq r,  \delta = \{1,2,...nv\}, j \neq i\}$
                \State
                \State \text{where $\delta$ is the set of nodes or robots in the network}
            \end{algorithmic}
        \end{algorithm}
    \end{minipage}
    \vspace{-20px}
\end{wrapfigure}

 %Zone load for robot $i$ ($L_i$) is calculated from equation~\ref{L_{pz}}, zone \textit{z} includes not only the workstations in the primary zone but also the transfer stations between each neighbor. Flow is only calculated from workstations in \textit{z} since each robot is only aware of the parts that are in its queue. This means that robots only know where to pick up and drop off parts next, not necessarily the starting pick-up and final drop-off locations where parts will be processed. If the robots only considered workstations that are in their primary zone, then parts that are at, or going to transfer stations will be missed when calculating zone load. 

Unlike GA and SA~\cite{ho_zone_2009}, DDZ cannot use part history to create new zones. Robots are only aware of where to pick up and drop off parts next, not their initial and final destinations. Incorporating part history creates an error when calculating load. If another zone inherits a transfer station, the zone that lost the transfer station will still include the transfer parts in its load calculation even though those transferred parts will no longer be running through that zone. 

Since robots are only aware of neighboring primary zones, every zone must be able to connect to at least one neighbor. Otherwise, parts leaving the zone will not be able to find a transfer station to exit through. For this reason, every zone must have available tip workstations for use as transfer stations. This means that when a zone connects to a transfer station, the connecting path is not considered when connecting to other transfer stations. This allows the connecting paths to cross each other but does not permit them to cross any other primary zones not associated with the two connecting zones.

\subsubsection*{System Overview}

To accompany the DDZ algorithm, a new zone management system is introduced to monitor, evaluate, and start zone redesign. Figure~\ref{sysoverview} shows a general overview of how each zone is handled along with how part and zone information is passed between the workstations and the robot. The system on every robot is divided into three main threads. The first is the control thread where zone and part data is sent out and received. The second is the monitoring thread where the robot runs Weighted Average Consensus~\cite{xiao_scheme_2005} periodically. The third is the correction thread where DDZ is initiated if the zone has been out of tolerance ($L_{TOL}$) for $T_{LT}$ time. 

The control thread is responsible for managing the robot queue and collecting data on neighbors. When a workstation sends out a part to be picked up, that part is then added to the robot queue where it will be scored and ranked with the rest of the parts in the queue using Shortest Job First with Aging (algorithm~\ref{shortest job}). The robot will then select the next part in the queue and move to pick up and drop off. Once the part is dropped off, it sends the part information to the workstation where it will be processed. When a part is selected, its information pops off of the queue and is stored. This is done to protect the part from getting lost if DDZ is started, ensuring that a new zone design will not take effect until the current selected part is dropped off. The control thread is also responsible for identifying neighbors and collecting data such as zone workstations and load. This information is later used in the monitoring and correction threads. In application, this thread is a combination of smaller threads that come together to perform the tasks described. 

The monitoring thread periodically runs Weighted Average Consensus~\cite{xiao_scheme_2005} (WAC) every $T_{AC}$ time. When it is time to start WAC, it sends a signal to neighboring nodes to also start WAC. Robot movement is preserved while running, updating the connected neighbors and recording their measured average load. Once WAC is finished, the measured average load ($x_i$) is sent to the correction thread. 

The correction thread uses $x_i$ and compares it to $L_i$. If the robot load has been out of tolerance $L_{TOL}$ for $T_{LT}$ minutes, then it will send a message to neighboring robots to start DDZ. The message is then propagated to other robots and continues during the duration of DDZ. If a robot does not have any neighbors, it continues normal operation until a neighbor is reached. Once DDZ is started, a robot will be declared a leader and continue with DDZ~\ref{DD_zoning}. To exit the algorithm, a robot must wait until all robots are done with DDZ. To do this, a robot monitors the status of neighboring robots. If all neighboring robots are finished, then the robot will exit and continue normal operations with the new zone design.  
\noindent\begin{minipage}{.5\textwidth}
    \begin{algorithm}[H]
    \caption{Decentralized Dynamic Zoning}\label{DD_zoning}
    \SetAlgoLined
        \SetKwInOut{Input}{Input}
        \SetKwInOut{Output}{Output}
        \Input{map of warehouse floor, number of zones, total number of episodes ($E_t$), total number of iterations ($I_t$), current workstation flow}
        \Output{zone design for robot $bestP_i$}
        \For{$ep \in E_t$}{
            wait until robot $i$ is selected as leader\\
            $N_i \gets$ set of neighbors belonging to $i$\\
            \textit{best}$\sigma_i \gets $standard deviation eq.~\ref{DDZstd}\\
            \textit{best}$P_i \gets$ best current zone design as seen from robot $i$\\ 
            \For{$n \in I_t$}{
                $P_i \gets$ current zone design\\
                $\sigma_i\gets$ standard deviation eq.~\ref{DDZstd}\\
                $j \gets$ random neighbor from $N_i$\\
                \eIf{$L_i >= L_j$}{
                    $i$ gives a random tip WS to $j$\\
                }{
                    $j$ gives a random tip WS to $i$\\
                }
                $P'_i \gets$ new zone design\\
                $P'_i \gets$ find new ts of $P'_i$ between all zones in $N_i$ and $i$ using algorithm~\ref{find_ts}\\
                re-queue every part in $N_i$ and $i$ using $P'_i$\\
                calculate new loads of $i$ and $j$ $\in N_i$ using $P'_i$\\
                $\sigma'_i\gets $standard deviation eq.~\ref{DDZstd} using $P'$\\
                $E \gets \sigma_i - \sigma'_i$\\
                $T_c \gets \textit{calc temp using eq. \ref{DDZtemp}}$\\
                $p \gets e^{\frac{E}{kT}}$\\
                $r \gets$ random number between 0 and 1\\
                \eIf{$\sigma'_i <= \sigma_i \vee r <= p$}{
                     $P \gets P'$\\
                }{
                    re-queue every part in $N_i$ and $i$ using $P_i$\\
                    calculate loads of $i$ and $j$ $\in N_i$ using $P_i$\\
                }
                \If{$\sigma'_i < \textit{best}\sigma_i$}{
                     $\textit{best}P_i \gets P'_i$\\
                     $\textit{best}\sigma_i \gets \sigma'_i$\\
                }
            }
         $P_i\gets \textit{best}P_i$\\
         select new robot as leader from $Ni$\\
        }
    wait for all robots in $N_i$ to finish before exiting\\
    \end{algorithm}
\end{minipage}%
\begin{minipage}[t]{.5\textwidth}
  \centering
  \vspace{-60ex}
  %\rule{0.3\textwidth}{50pt}
  \includegraphics[width=3.48in]{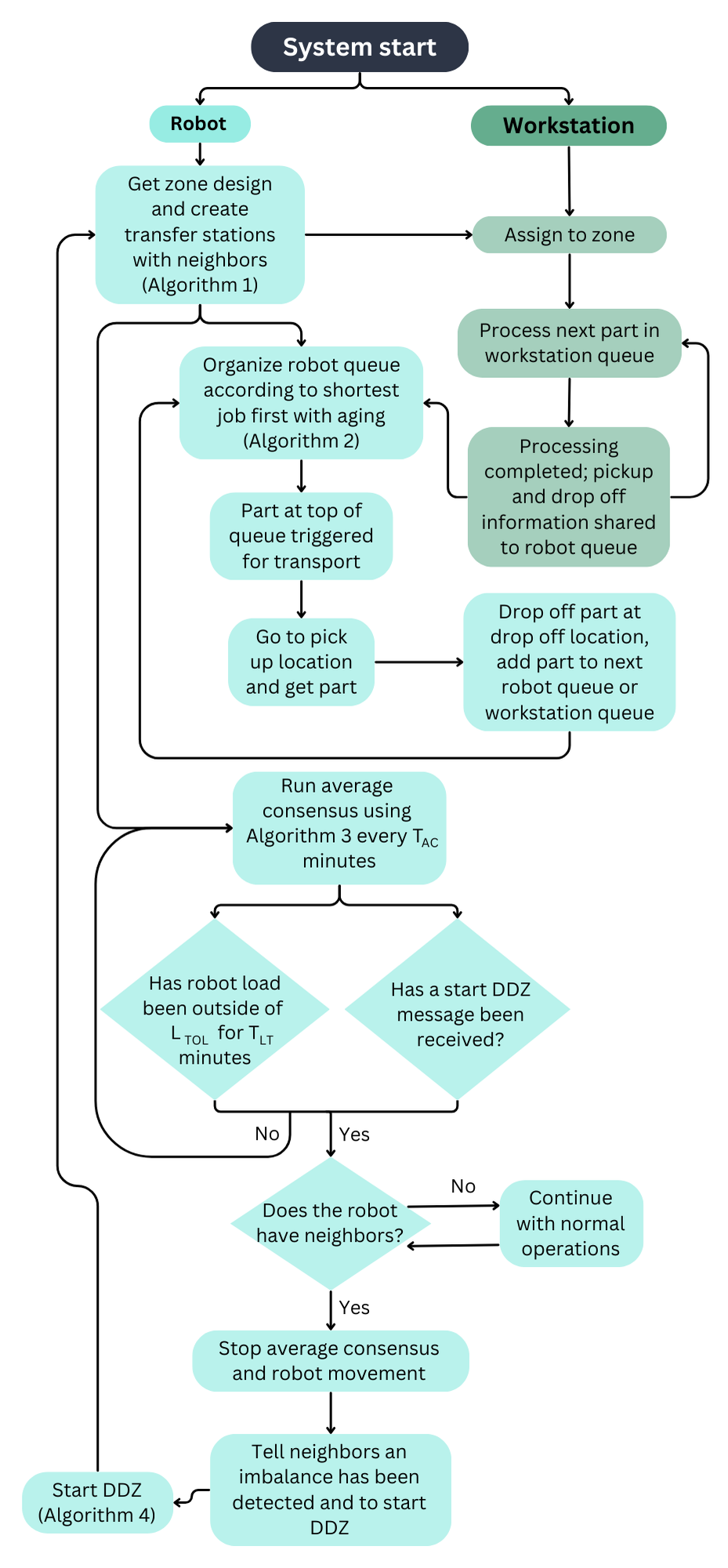}%
  \captionof{figure}{DDZ system flowchart} 
  \label{sysoverview}
\end{minipage}
\section*{Results}

%Up to three levels of \textbf{subheading} are permitted. Subheadings should not be %numbered.

\subsection*{Simulation} \label{simulation}
\begin{comment}
\begin{wrapfigure}[16]{r}{0.5\textwidth}
    \centering
    \includegraphics[width=3.5in]{images/manf_floor.jpg}
    \caption{Simulated manufacturing floor; yellow: shipping and receiving, blue: warehouse, red: production/assembly area }
    \label{floor}
\end{wrapfigure}

\begin{wrapfigure}[16]{r}{0.5\textwidth}
    \centering
    \includegraphics[width=1.52in]{images/LElayout_paper.JPG}
    \caption{Simulation floor layout (dimensions are in feet)}
    \label{sim layout}
\end{wrapfigure}
\end{comment}

\begin{figure}[!b]
\centering
\begin{subfigure}{.45\textwidth}
  \centering
  \includegraphics[width=3.0in]{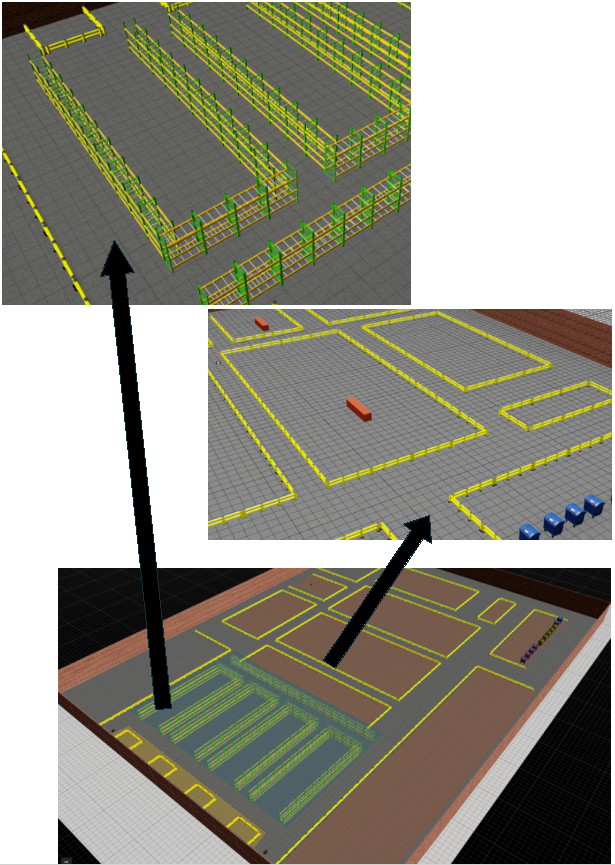}
    \caption{Simulated manufacturing floor; yellow: shipping and receiving, blue: warehouse, red: production/assembly area }
  \label{floor}
\end{subfigure}%
\begin{subfigure}{.5\textwidth}
  \centering
  \includegraphics[width=3.7in]{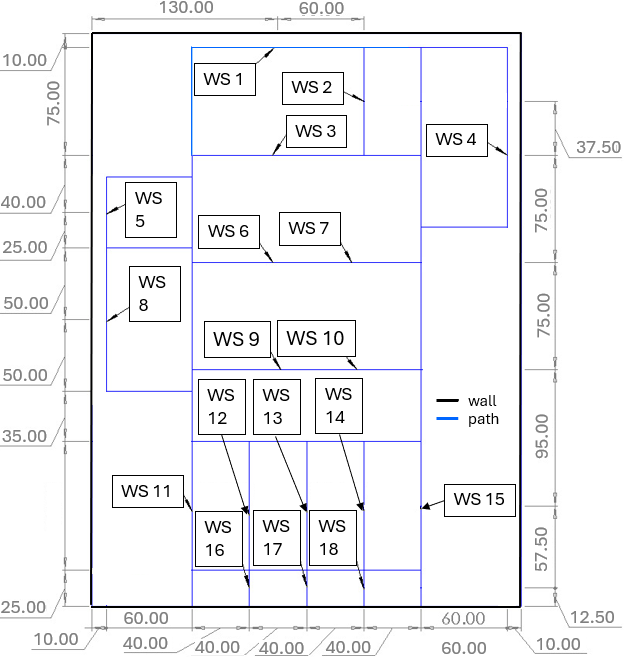}
  \caption{This image gives context and scale to the manufacturing floor used in the simulation as well as the workstation (WS) location (dimensions are in feet)}
  \label{floormap}
\end{subfigure}
\caption{Figure~\ref{floor} and figure~\ref{floormap} shows the environment and scale of the simulation}
\label{env}
\end{figure}

The simulation tests how each method performs under realistic conditions. The purpose is to show how DDZ adapts to a dynamic work environment and compare them to GA and SA from Ho and Liao~\cite{ho_zone_2009}. The performance metrics used are throughput and distance traveled. The simulation is run with three AMRs in a simulated manufacturing floor with a section of it representing a warehouse and others representing assembly/production locations (see figure~\ref{floor}). Figure~\ref{floormap} shows the workstation (WS) locations on the map. Each algorithm was trained on a set of parts to find an initial zone design. Then when the simulation started, a different set of parts with alternate routes was run through the system. This forced the method to find a better zone design once an imbalance was detected.

The SA~\cite{ho_zone_2009} algorithm was modified to adjust it for use with AMRs. Transfer stations were found using algorithm~\ref{find_ts}. Algorithm~\ref{find_ts} bypasses the need to assign transfer stations based on how much of the connecting path is inside either zone. The other change is how load sharing is performed. In this paper, load sharing does not consider the other vehicles' location or path when delivering a part into the next zone. GA operates in the same manner as the SA algorithm except GA does not need to find an initial zone design. Instead, GA finds the optimal zone by iteratively evaluating the balance of a large solution set of different zones. 

\begin{minipage}{.49\linewidth}
    \centering
    \begin{tabular}{|c|c|c|}\hline
        \textbf{Part} & \textbf{Route} & \textbf{Qty}\\
        \hline
        A & 4,2,1,3,14 & 30\\
        B & 5,6,7,13,17 & 30\\
        C & 11,8,9,10,18 & 20\\
        D & 14,11,13,11,8,10,14,15 & 20\\
        \hline
    \end{tabular}
    \captionof{table}{Part types and their routes. Route numbers are workstations from figure~\ref{floormap} e.g. WS(4)}
    \label{routes2}
\end{minipage}
\begin{minipage}{.49\linewidth}
  \centering
    \begin{tabular}{|c|c|}\hline
        \textbf{Workstation} & \textbf{Time (min)}\\
        \hline
        WS1, WS5, WS15 & 6\\
        WS2, WS7, WS9 & 4\\
        WS3, WS6, WS8 & 3\\
        WS4, WS10, WS17, WS18 & 5\\
        WS11, WS12, WS13, WS14, WS16 & 1\\
        \hline
    \end{tabular}
    \captionof{table}{Part processing times}
    \label{processing time}
\end{minipage} 
\newline
\newline
The simulation was built with NVIDIA ISAAC Sim~\cite{nvidia_developer_isaac_2024} with a manufacturing floor layout as shown in figure~\ref{env}. The simulation used a 2.2 GHz CPU, NVIDIA GeForce RTX 3070, and 32 GB of RAM. Localization, path planning, and obstacle avoidance were handled through ROS Navigation. Each robot was an NVIDIA Carter~\cite{nvidia_developer_isaac_2024} robot with three wheels in a differential drive configuration. Sensors included a LiDAR camera and an IMU for localization. 

Each part type from table~\ref{routes2} was simultaneously run through the system. Parts go into a workstation queue where they wait to be processed according to table~\ref{processing time}. Once processed, parts are added to a robot's queue where they are picked up and taken to the next workstation. After zone repair, parts are reassigned to the corresponding robot and zone where the part is currently residing. For GA and SA~\cite{ho_zone_2009}, a 20-minute rolling window was used for data recording. This means that once a part has exited the robot queue, the history of that delivery will only be kept for 20 minutes. See~\cite{lab_aralab-unrdynamic-zoning_2024} for a list of all parameters used in the simulation.

\subsection*{Comparison}

%\begin{wrapfigure}{r}{0.5\textwidth}
\begin{figure}
    \centering
    \includegraphics[width=4.0in]{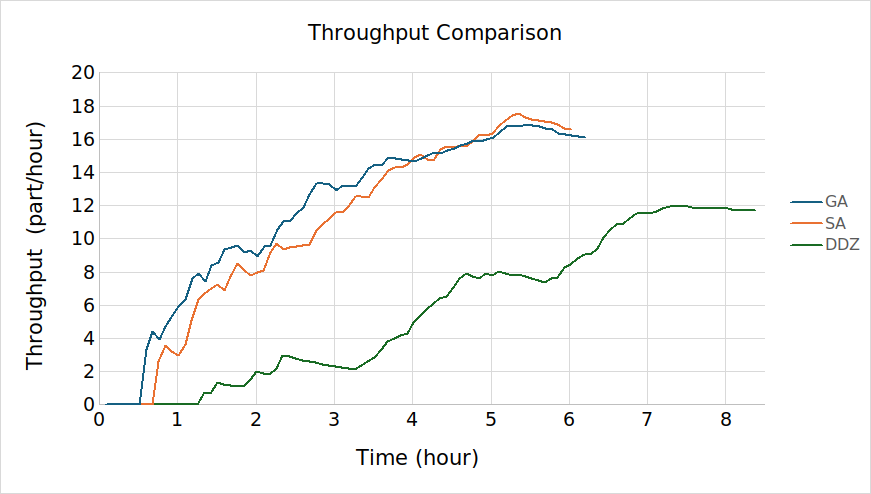}
    \caption{SA, GA, DDZ throughput comparison}
    \label{throughput}
%\end{wrapfigure}
\end{figure}

The results of the simulation are summarized in table~\ref{results} and figure~\ref{throughput}. Once robot queues were loaded with parts to deliver, DDZ detected an imbalance and started the timer for zone repair. Eventually, each algorithm moved into zone repair and found new zones. After zone repair, each system continually moved from balanced, to imbalanced, and then zone repair based on the current needs of the simulation until each part was finished. 
\begin{table}[t]
    \centering
    \begin{tabular}{|p{.45in}|p{.9in}|p{.9in}|p{.9in}|p{.9in}|p{.9in}|}\hline
        \textbf{Method} & \textbf{Time to complete (hour)} & \textbf{\% time in balance} &\textbf{Average travel dist (feet)} & \textbf{$\sigma$ travel dist (feet)}\\
        \hline
        SA & 5.94  & 39.32 &56220.83 & 4145.01\\
        GA & 6.13  & 23.71 &54454.17 & 3678.93\\
        DDZ & 8.62 & NA & 66250.67 & 1295.22\\
        \hline
    \end{tabular}
    \caption{Simulation results of SA, GA, and DDZ}
    \label{results}
\end{table}
Comparing throughput and total distance traveled, GA performed better than SA. However, SA was able to finish processing the 100 parts about 12 minutes faster than GA while remaining in balance \%39.32 of the time. GA had a lower average travel distance than SA with less total distance deviation between each AMR. This shows that GA was able to move each AMR a more equal distance than SA. DDZ in terms of performance, was the slowest taking about 3 more hours to finish processing the 100 parts. However, DDZ was shown to distribute the load more effectively than SA~\cite{ho_zone_2009} and GA. This is shown in the total travel distance deviation in table~\ref{results}. The discrepancy in throughput performance can be attributed to DDZ not being able to "load share" when an imbalance is detected. During "load share", SA~\cite{ho_zone_2009} and GA stop moving parts through transfer stations and instead directly deliver parts to their next processing station. For DDZ, to avoid the chance of robots clumping together, potentially leading to zone isolation, the robots operate relatively close to their designated zone and only allow parts to transfer into a new zone through transfer stations.

\subsubsection*{Code availability}
The simulation, ROS packages, and map for the simulation can be found here~\cite{lab_aralab-unrdynamic-zoning_2024}.

%\subsection*{Subsection}

%Example text under a subsection. Bulleted lists may be used where appropriate, e.g.

%\begin{itemize}
%\item First item
%\item Second item
%\end{itemize}

%\subsubsection*{Third-level section}
 
%Topical subheadings are allowed.

\section*{Discussion}

%The Discussion should be succinct and must not contain subheadings.

In this paper, we build upon the zone design method first presented in Ho and Liao~\cite{ho_zone_2009} and introduce a new decentralized algorithm for use with AMRs in manufacturing and warehouse environments. To summarize, zones are built by connecting workstations by critical segments. The workload balance of these zones is evaluated by measuring the operational time of each robot in a zone. Workload balance is achieved by adjusting each zone so that robots will share a similar operational time. 

DDZ eliminates the need for a central computer to find an optimal zone design. The algorithm starts by finding the average load of the entire system through Weighted Average Consensus with Metropolis Weights~\cite{xiao_scheme_2005}. Then each robot takes turns finding an optimal zone design between each of its neighbors by running to search for a zone design with the lowest standard deviation of load.  

Preliminary results show GA can have a similar throughput to that of SA while distributing the load more equally among the robots. This is demonstrated by GA having a smaller variance in total travel distance between each robot. DDZ had a worse throughput performance but was shown to have the smallest variance in total travel distance between each robot when compared to SA and GA. The reduced throughput performance can be attributed to DDZ not being able to perform load sharing when an imbalance is detected. Instead, DDZ forces every part to go through a transfer station despite the status of the system.

%It is important to note that the standard deviation of the total travel distance is only a snapshot of the total load of the robot. During the simulation, the distances traveled by each robot vary depending on the current status of the system. It is also controlled by how small the tolerance is for the zone to be declared balanced. A smaller tolerance could see a smaller distribution of load across each robot operating in the system.

\section*{Conclusion} \label{conclusion}
%Autonomous mobile robots (AMRs) have significantly contributed to manufacturing, particularly in warehousing, by enhancing performance and productivity. Advances in localization, sensors, and battery management have made AMRs more feasible for deployment, while AI techniques have become crucial in the decision-making process of scheduling. With the development of zoning presented in this paper, AMRs have shared the workload of laborious and repetitive tasks allowing humans to be less fatigued and focused. Further development of zoning will continue to increase productivity while growing the human-robot collaboration space. 
In conclusion, this paper introduces a novel Decentralized Dynamic Zoning (DDZ) algorithm tailored for autonomous mobile robots (AMRs) in industrial environments. The proposed DDZ approach enhances task distribution and workload balancing by dynamically adjusting zones based on real-time conditions. Unlike previous centralized methods, DDZ leverages local interactions between robots to achieve a more balanced distribution of tasks, thereby reducing travel distance variation and supporting efficient operations in unpredictable settings. While DDZ demonstrated a lower throughput in simulations compared to other methods, it successfully minimized total travel distance deviation among AMRs, indicating its potential for consistent task allocation in real-world applications. This decentralized approach opens new pathways for efficient AMR management, particularly in complex manufacturing and warehouse environments.
\section*{Future work} \label{future work}

%Although the GA works well for the scenario presented in this paper, it will be beneficial to explore different heuristic algorithms. An example could be the use of particle swarm optimization which uses agents to explore a solution space. Each agent moves toward an improved solution, receiving influence both from neighbors and the collective. Being another heuristic approach, it is expected to perform similarly to SA and GA. 

%Currently, parts are added to robots through a sorted queue. This method of scheduling within the zone is inefficient, as it overlooks parts that could be dropped off en route to pick up another. Alternative scheduling algorithms, such as that of Saylam \emph{et al.}~\cite{saylam_minmax_2023}, can help schedule parts to move through each zone more efficiently. 

Though this proposed algorithm is meant for use with AMRs, which can handle path planning and obstacle avoidance, this paper doesn't consider other factors like charging time or robot malfunction. Currently, DDZ relies on assuming that every robot runs at the same speed and that robots do not need to charge. In a real-world scenario, this system will need to consider the charging schedules of each robot and enable another robot to take over a zone if one becomes incapacitated. 

DDZ also assumes that every AMR is the same type and can have the same weight. Sectioning off areas of the floor can also have the benefit of having different robot types occupy their specialized zone. This could be an addition to DDZ by assigning robot priority to certain zones while preventing other AMRs with different capabilities from entering. 

DDZ's poor throughput performance can be improved by implementing an operational mode like that of load sharing. Robots would still need to remain relatively close to their designated zones; however, if the final destination is near a transfer station, it would be beneficial to drop off at the next processing workstation.

\section*{Data availability}
All data generated or analyzed with open-source codes during this study are included in this published article. The simulation, ROS packages, map, and data from the simulation can be found here https://github.com/aralab-unr/Dynamic-zoning.git.

\bibliography{AMR_Libary2}

%\noindent LaTeX formats citations and references automatically using the bibliography records in your .bib file, which you can edit via the project menu. Use the cite command for an inline citation, e.g.  \cite{Hao:gidmaps:2014}.

%For data citations of datasets uploaded to e.g. \emph{figshare}, please use the \verb|howpublished| option in the bib entry to specify the platform and the link, as in the \verb|Hao:gidmaps:2014| example in the sample bibliography file.

%\section*{Acknowledgements (not compulsory)}

%Acknowledgements should be brief, and should not include thanks to anonymous referees and editors, or effusive comments. Grant or contribution numbers may be acknowledged.

\section*{Author contributions statement}
 
R.K. designed the methods described above and built the simulation. R.K. also conducted the simulation and analyzed the results. H.L. provided research guidance and editorial input for this article. All authors critically reviewed the manuscript.

\section*{Additional information}\label{addinfo}

%#To include, in this order: \textbf{Accession codes} (where applicable); 
The authors declare no competing interests.

%The corresponding author is responsible for submitting a \href{http://www.nature.com/srep/policies/index.html#competing}{competing interests statement} on behalf of all authors of the paper. This statement must be included in the submitted article file.

%\begin{figure}[ht]
%\centering
%\includegraphics[width=\linewidth]{stream}
%\caption{Legend (350 words max). Example legend text.}
%\label{fig:stream}
%\end{figure}

%\begin{table}[ht]
%\centering
%\begin{tabular}{|l|l|l|}
%\hline
%Condition & n & p \\
%\hline
%A & 5 & 0.1 \\
%\hline
%B & 10 & 0.01 \\
%\hline
%\end{tabular}
%\caption{\label{tab:example}Legend (350 words max). Example legend text.}
%\end{table}

%Figures and tables can be referenced in LaTeX using the ref command, e.g. Figure \ref{fig:stream} and Table \ref{tab:example}.

\begin{comment}
\begin{wrapfigure}{r}{0.5\textwidth}
    \begin{minipage}{0.5\textwidth}
        \begin{algorithm}[H]
        \caption{Weighted Average Consensus with Metropolis Weights}\label{WAC}
            \begin{algorithmic}
                \State $x_i(t+1)=W_{ii}(t)x_i(t)+\sum_{j \in N_i(t)}W_{ij}(t)x_j(t)$,   $i = 1, ...,nv$
                \State \[ W_{ij} = \begin{cases}
                    \frac{1}{1+\max\{d_i(t),d_j(t)\}},& \text{if} \{i,j\}\in\xi(t)\\
                    1-\sum_{\{i,k\}\in\xi(t)}W_{ik}(t), & i=j\\
                    0, & \textit{otherwise}
                \end{cases}
                \]
            \end{algorithmic}
        \end{algorithm}
    \end{minipage}
\end{wrapfigure}
\end{comment}

\end{document}